\documentclass[runningheads]{llncs}
\usepackage{graphicx}

\usepackage[linesnumbered,ruled]{algorithm2e}
\usepackage{subcaption}
\usepackage{paralist}
\usepackage{amsmath}
\usepackage{amssymb}
\usepackage{xcolor}
\usepackage{orcidlink}
\usepackage[capitalise]{cleveref}

\crefname{algorithm}{Alg.}{Algs.}
\Crefname{algorithm}{Algorithm}{Algorithms}

\begin{document}
\title{A First Step Towards Even More Sparse Encodings of Probability Distributions\thanks{Final publication in Katzouris, N., Artikis, A. (eds) Inductive Logic Programming. ILP 2021. Lecture Notes in Computer Science, vol 13191. Springer, Cham. The final authenticated publication is available online at \url{https://doi.org/10.1007/978-3-030-97454-1_13}}}
\titlerunning{Towards Even More Sparse Encodings of Probability Distributions}
\author{Florian Andreas Marwitz\orcidlink{0000-0002-9683-5250} \and
Tanya Braun\orcidlink{0000-0003-0282-4284} \and Ralf Möller\orcidlink{0000-0002-1174-3323}}
\authorrunning{F. A. Marwitz et al.}
\institute{Institute of Information Systems, University of Lübeck, Lübeck, Germany\\
\email{\{florian.marwitz@student,braun@ifis,moeller@ifis\}.uni-luebeck.de}}
\maketitle              %
\begin{abstract}
Real world scenarios can be captured with lifted probability distributions. 
However, distributions are usually encoded in a table or list, requiring an exponential number of values. 
Hence, we propose a method for extracting first-order formulas from probability distributions that require significantly less values by reducing the number of values in a distribution and then extracting, for each value, a logical formula to be further minimized.
This reduction and minimization allows for increasing the sparsity in the encoding while also generalizing a given distribution.
Our evaluation shows that sparsity can increase immensely by extracting a small set of short formulas while preserving core information.

\end{abstract}

\section{Introduction}

Modeling real world scenarios requires dealing with uncertainties.
A full joint probability distribution, factorized into local distributions for a sparse encoding, over a set of random variables (randvars) allows for modeling such scenarios.
With first-order logic, we can compactly encode relationships between large sets of randvars, representing sets of indistinguishable randvars by parameterizing randvars with logical variables (logvars).
However, encoding local distributions over (parameterized) randvars usually relies on the values stored in a table or list for ease of handling the encoding, with more compact encodings like algebraic decision diagrams leading to a huge overhead \cite{ChaDa07}.
Thus, there is an exponential number of values to store per local distribution, also called factor or, if logvars are involved, parfactor.

Turning to first-order logic for a sparse encoding, Markov Logic Networks (MLNs)~\cite{mln} use weighted first-order logic (FOL) formulas to represent a probability distribution compactly.
Canonically transforming a parfactor into formulas translates each entry in the parfactor into one formula (given Boolean ranges of the randvars), which means an exponential number of formulas~\cite{mln_to_parfactor}.
Therefore, this paper works towards an even more sparse encoding by reducing the number of values in a distribution, allowing for combining different entries into a single formula.
Specifically, this article presents CoFE (\underline{Co}mpact \underline{F}ormula \underline{E}xtraction), a method for extracting FOL formulas from parfactors.
We test out two strategies for reducing the number of values in a parfactor, guided by an $\epsilon$ margin that caps the distance between the original distribution and the modified distribution.
Our proof-of-concept evaluation shows that CoFE makes a reduction in the number of formulas possible and can even be robust against noise added to values, while keeping the error in query answers in the reduced model small.

The theoretical foundations for query answering in parfactor graphs are laid by Poole with first-order probabilistic inference~\cite{fo_inference}, also introducing parfactors as a modeling formalism.
Richardson and Domingos introduce MLNs as another approach to combine FOL and probabilistic graphical models~\cite{mln}.
There exist various query answering algorithms, exact and approximate, that work with either parfactors or MLNs, e.g., \cite{smokers,GogDo11,AhmKeMlNa13,BraMo16a}.
In terms of related work, there exist well-established techniques in statistics to approximate a discrete randvar with another discrete or continuous randvar.
Please refer to~\cite{chakraborty2015generating} for details. 
However, the problem these techniques solve does not apply here as it lacks the first-order aspect of MLNs and parfactors.
There exists a range of probabilistic logic learners that return a set of weighted FOL formulas, of which ProbLog~\cite{RaeKiTo07} is a prominent representative.
However, again, the problem setting does not apply as these learners have a set of (positive and negative) samples, which is not available in our case.
Statistical relational learners such as the boosted tree learner \cite{NatKhKeGuSh12} also focus on the problem of learning a model from a set of samples.
We have a model in the form of a set of parfactors given, which we want to transform into an MLN to preserve semantics while reducing the number of formulas necessary.

The rest of this paper is structured as follows: 
First, we define and explore the required math. 
Second, we present CoFE. 
Third, we evaluate CoFE empirically. 
Last, we end with a conclusion.

\section{Notations and Problem Statement}
In this section, we define parfactors and MLNs. 
Parfactors are functions mapping argument values to real numbers called potentials. 
An MLN is a set of pairs of a FOL formula and a weight. 
Furthermore, we show the transformation of a parfactor to an MLN and define a distance for two parfactors. 
Definitions for parfactors are mainly based on~\cite{parfactor_definitions} and for MLNs on~\cite{mln}.

\begin{definition}[Parfactor model]\label{def:factor}
	Let $\mathbf{R}$ be a set of randvar names, $\mathbf{L}$ a set of logical variable (logvar) names, $\Phi$ a set of factor names, and $\mathbf{D}$ a set of constants (universe).
	All sets are finite.
	Each logvar $L$ has a domain $\mathcal{D}(L) \subseteq \mathbf{D}$.
	A \emph{constraint} $C$ is a tuple $(\mathcal{X}, C_{\mathcal{X}})$ of a sequence of logvars $\mathcal{X} = (X_1, \dots, X_n)$ and a set $C_{\mathcal{X}} \subseteq \times_{i = 1}^n\mathcal{D}(X_i)$.
	The symbol $\top$ for $C$ marks that no restrictions apply, i.e., $C_{\mathcal{X}} = \times_{i = 1}^n\mathcal{D}(X_i)$.
	A parameterized randvar \emph{(PRV)} $R(L_1, \dots, L_n), n \geq 0,$ is a syntactical construct of a randvar $R \in \mathbf{R}$ possibly combined with logvars $L_1, \dots, L_n \in \mathbf{L}$. %
	If $n = 0$, the PRV is parameterless and constitutes a propositional randvar.
	The term $\mathcal{R}(A)$ denotes the possible values (range) of a PRV $A$. 
	An \emph{event} $A = a$ denotes the occurrence of PRV $A$ with range value $a \in \mathcal{R}(A)$. 
	We denote a \emph{parfactor} $g$ by $\phi(\mathcal{A})_{| C}$ with $\mathcal{A} = (A_1, \dots, A_n)$ a sequence of PRVs, $\phi : \times_{i = 1}^n \mathcal{R}(A_i) \mapsto \mathbb{R}^+$ a \emph{potential function} with name $\phi \in \Phi$, and $C$ a constraint on the logvars of $\mathcal{A}$.
	A set of parfactors forms a \emph{model} $G := \{g_i\}_{i=1}^n$.
	With $Z$ as normalizing constant, $G$ represents the full joint distribution $P_G = \frac{1}{Z} \prod_{f \in gr(G)} f$, with $gr(G)$ referring to the groundings of $G$ w.r.t. given constraints.
\end{definition}
\emph{Parfactor size} refers to the size of the range of the parfactor arguments $\mathcal{A}$, i.e., $|\mathcal{R}(\mathcal{A})|$. 
Parfactors only contain universal quantifiers.
For comparing parfactors, we need to define a distance. 
In this paper, we use the Hellinger distance, defined for probability distributions, as it allows to have zeroes in the distribution.
One could however use any distance function of their own choosing.
As the potentials in a parfactor do not need to form a probability distribution, we normalize the potentials for calculating the Hellinger distance between two parfactors.
\begin{definition}[Hellinger distance]
Let $\phi_1(\mathcal{A})$ and $\phi_2(\mathcal{A})$ be two parfactors defined over the same PRVs $\mathcal{A}$. 
Let $\Sigma_1$ and $\Sigma_2$ denote the sum of the potentials of $\phi_1$ and $\phi_2$, respectively.
The \emph{Hellinger distance} is then defined as
\begin{align*}
H(\phi_1(\mathcal{A}), \phi_2(\mathcal{A})) = \frac{1}{\sqrt{2}} \sqrt{\sum_{\boldsymbol{a}\in \mathcal{R}(\mathcal{A})}\left( \sqrt{\frac{\phi_1(\boldsymbol{a})}{\Sigma_1}} - \sqrt{\frac{\phi_2(\boldsymbol{a})}{\Sigma_2}} \right)^2}.
\end{align*}
\end{definition}

\begin{definition}[MLN]
An \emph{MLN} $M$ is a set of pairs $(F_i,w_i)$, where $F_i$ is an FOL formula and $w_i \in \mathbb{R}$.
	With $Z$ as normalizing constant, $M$ represents the full joint distribution $P_M = \bigcup_{\boldsymbol{x} \in \mathcal{R}(\boldsymbol{X})}\frac{1}{Z} \exp \left( \sum_i w_i n_i(\boldsymbol{x}) \right)$, where $\boldsymbol{X}$ is the set of all grounded randvars in $M$, $w_i$ the weight of formula $F_i$, and $n_i(x)$ the number of true groundings of $F_i$ in $\boldsymbol{x}$.
\end{definition}

\subsubsection{The Problem}
We can canonically transform a parfactor into an MLN by adding a formula for every potential given Boolean ranges of PRVs. 
For non-boolean PRVs, transforming parfactors is more elaborate.
Due to the exponential function in the semantics of an MLN, the weight is the natural logarithm of the potential. 
Consider the parfactor $\psi(Friends(X,Y), Smokes(X), Smokes(Y))$ from the smokers dataset~\cite{smokers}, %
which maps $(1, 1, 1)$ to the potential $7.39$ and the remaining range value combinations to the potential $1$.
We add a formula for each range value combination of the three PRVs together with the natural logarithm of the potential: 
For $\psi(1,1,1)=7.39$, we add the pair $(a \land b \land c, 2)$, with $a,b,c$ referring to the three PRVs being set to true. 
For $\psi(1,1,0)=1$, we add the pair $(a \land b \land \lnot c, 0)$ and so on. 
Thus, we have as many formulas as is the parfactor size. 
But instead of eight formulas, we would like to extract only two:
\begin{align}
0 &\quad \lnot friends(X, Y) \lor \lnot smokes(X) \lor \lnot smokes(Y) \label{eq:smokers_rule1}\\
2 &\quad friends(X, Y) \land smokes(X) \land smokes(Y) \label{eq:smokers_rule2}
\end{align}
which we could even reduce to one formula given the MLN semantics and the fact that formulas that evaluate to false receive the weight $0$.

The smokers example showcases the power of compactly encoding a distribution with few formulas.
However, the potentials are rarely distributed this nicely.
Consider the example parfactor in the first four columns of \cref{tab:simplification}. 
With the canonical transformation, we transform the parfactor into eight formulas. 
But if we reduce the number of different potentials, a single formula can encode more than one potential. 
If we map, for example, lines 2 to 7 to the potential $5$, reducing the number of different potentials from $8$ to $2$, we can encode the same information in two formulas, $(\lnot a \land \lnot b \land \lnot c, \ln 1)$ and $(a \lor b \lor c, \ln 5)$. 
In this example, we have an exponential reduction in the amount of extracted formulas. 

\begin{table}[tb]
\centering
\caption{Reduction result for a parfactor $\phi : \{0,1\}^3 \to \mathbb{R}^+$. 
The last two columns show which numbers are mapped to the same one when applying the respective strategy. 
DBSCAN parameters are $\theta_d = 1, \theta_n = 1$. 
Without reduction, we need eight formulas. With quartiles, we need four formulas and two with clustering.}
\label{tab:simplification}
\begin{tabular}{ccc c cc}
$a \in \{0,1\}$ & $b \in \{0,1\}$ & $c \in \{0,1\}$ & $\phi(a,b,c)$ & quartile & cluster \\
\hline
0 & 0 & 0 & 1   & 1 & 1 \\
0 & 0 & 1 & 4.7 & 1 & 2 \\
0 & 1 & 0 & 4.8 & 2 & 2 \\
0 & 1 & 1 & 4.9 & 2 & 2 \\
1 & 0 & 0 & 5   & 3 & 2 \\
1 & 0 & 1 & 5.1 & 3 & 2 \\
1 & 1 & 0 & 5.2 & 4 & 2 \\
1 & 1 & 1 & 5.3 & 4 & 2 \\
\end{tabular}
\end{table}

\section{CoFE: Compact Formula Extraction} \label{sec:extraction}
As we argue above, directly transforming parfactors to formulas still requires an exponential number of values. 
Therefore, we propose CoFE for extracting compact formulas from parfactors by reducing the number of different potentials in a parfactor.
The formula extraction process consists of three steps,
\begin{inparaenum}[(1)]
	\item \emph{reduction}, 
	\item \emph{extraction}, and 
	\item \emph{minimization}.
\end{inparaenum}
The individual steps are explained below.

The first step is \emph{reduction}.
The goal is to reduce the amount of different numbers in a parfactor while modifying the distribution only minimally. 
We formalize the notion of minimal modification by considering the distance between the original and the modified version, which should be lower than a predefined maximum distance $\epsilon$.
For reduction, we test out two straight-forward strategies, based on quantiles and clustering, respectively.
Investigating more complex strategies is left as future work.
In the \emph{quantile} strategy, we calculate $q$-quantiles and map each number belonging to a quantile to the mean of the quantile. 
We increase $q$ ($q=1,2,3, \ldots, \text{parfactor size}-1$) until the distance is smaller or equal to $\epsilon$. 
In the worst case, we do not modify the potential function at all as all $q$s might yield a distance larger than $\epsilon$.
In the \emph{clustering} strategy, we cluster the numbers in the potentials and map each number of a cluster to the mean of the cluster. 
For clustering, we use DBSCAN~\cite{dbscan,schubert2017dbscan}, a density-based clustering algorithm, as it does not need the amount of clusters ($k$) as input. 
It needs two other parameters, though: a threshold distance between two points to be considered neighbors~($\theta_d$) and a minimum number of neighbors to be classified as a cluster~($\theta_n$), which are user-defined.
Other clustering methods can be used as well. 
If such a method requires a $k$ as an input, a similar approach can be followed as for the quantile strategy, starting with $k=1$ and increasing $k$ up until the parfactor size minus one.
Note that both strategies use the mean as mapping target because the distance between numbers and their mean is short.
An example for both strategies is given in \cref{tab:simplification}, showing which cluster or quantile ($q=4$, i.e., quartiles) an entry is mapped to. 
The quantile strategy maps to four numbers.
Clustering makes use of the accumulation around the number five and maps to only two distinct numbers.
In the implementation for the evaluation, CoFE actually chooses the result of the strategy that maps to fewer different numbers, breaking ties by lower distance.

Next, the \emph{extraction} step follows:
CoFE extracts logical formulas in the same way as a parfactor is transformed into an MLN, getting a list of formulas, each assigned with a weight. 
The third step is \emph{minimization}, for which CoFE sorts the formulas sharing the same weight into buckets labeled with this weight. 
Then, the formulas of each bucket are set to be minimized into one minimal formula, for which CoFE uses the Quine-McCluskey algorithm~\cite{qmc}.
The output for a parfactor is the set of minimized logical formulas, each assigned with a weight.
Consider the smokers example again:
When we use the parfactor $\psi$, as described above, as input for CoFE with parameters $\epsilon = 0.1, \theta_d = 0.1, \theta_n = 1$, CoFE correctly returns the two formulas given in \cref{eq:smokers_rule1,eq:smokers_rule2}.
We could also apply CoFE to answers to queries for (conditional) probability distributions over a set of randvars, turning answers into formulas as well.

\subsubsection{On the $\epsilon$ and Its Effect on Error and Reduction}
If applying CoFE to each parfactor in a parfactor model and unifying the outputs, we get an MLN representing the same full joint distribution as the set of reduced parfactors.
If using this MLN (or the reduced parfactor model) for query answering, then the query results can diverge from the result of the original model.
As mentioned before, CoFE relies on a user-defined $\epsilon$.
With a large $\epsilon$, CoFE is able to reduce more potentials but we expect the divergence in query results to rise.
With a small $\epsilon$, CoFE most likely reduces fewer numbers while we expect query results to not diverge to a large extent.
In a worst case scenario, CoFE is not able to reduce the number of formulas at all, i.e., there are $r^n$ formulas for the $r^n$ potentials the canonical table representation has, with $r$ being the range cardinality of the $n$ PRVs the parfactor is defined over.
However, with a large enough $\epsilon$ and an optimal minimization result, we get $k$ clusters or $q$ quantiles leading to $k$ or $q$ formulas of length $n$, which is no longer exponential in $n$.
The upcoming evaluation looks into both reduction and errors in query answering empirically.

\section{Empirical Evaluation}
In this section, we evaluate CoFE empirically.
First, we describe the test setting in more detail. 
Second, we look at CoFE's capability of reconstructing a given formula, distorting potentials by adding noise. 
Third, we evaluate the number and length of formulas CoFE outputs as well as the error it incurs during query answering and briefly discuss the results in the light of readability.

\subsection{Test Setting}
The first part of the evaluation looks at CoFE's performance reconstructing latent formulas under noise:
By applying CoFE to a dataset in which noise is added to potentials, we can show that the algorithm is robust against noise. 
In particular, we can simulate extracting latent formulas from a dataset by treating the original model as the latent one and the noisy model as the given input.
We may not add too much noise, because otherwise the noisy model would lose information and CoFE could not extract anything useful.

The second part looks at the sparsity as well as the incurred error.
A reduced number of formulas comes along with deviations in query answers if using the extracted formulas for inference. 
Thus, we take a look at the number of formulas and their length as well as the mean error for queries in the reduced model, which also indicates how much information encoded in a distribution we preserve.
As queries, we use a set of representative queries where for each PRV in a model we pose a query with an arbitrary grounding.
The error is the deviation of the query answer on the mapped model from the query answer on the noised model.

We perform two tests on the smokers example from~\cite{smokers}, which we have used throughout this article as an example. 
The tests differ in the noise standard deviation $\sigma$.
Moreover, we create an artificial dataset for investigating the effect of the ratio of the cluster sizes. 
We create a model with nine parfactors, each defined over three randvars. 
Parfactor $g_i, i=1 \dots 9,$ has $i-1$ ones and $9-i$ twos in the potentials. 
We perform two tests with different standard deviations for noise on the artificial dataset. 
\Cref{tab:test_parameters} shows the test parameters. We choose the parameters so that clustering is applicable and that clusters can span over a standard deviation of the added noise.

Specifically, we have the following steps for each test:
First, we add a normally distributed noise with a mean of zero to the potentials in parfactors. 
Next, we apply CoFE to the noised model. 
For the first part, we calculate the Hellinger distances between the original model, the noised, and the extracted one. 
For the second part, we look at the formulas and the mean error for query answering.

\begin{table}[tb]
\caption{Evaluating CoFE: Number of formulas extracted per parfactor ($\#$), number of atoms per formula ($L$), and the mean absolute error ($E$).}
\begin{subtable}[c]{0.5\textwidth}
\centering
\subcaption{Test parameters} \label{tab:test_parameters}
\begin{tabular}{c|ccccc}
test name & dataset & $\sigma$ & $\epsilon$ & $\theta_d$ & $\theta_n$\\
\hline
Smokers1 & smokers & 0.5 & 0.3 & 2 & 2 \\
Smokers2 & smokers & 1 & 0.3 & 2 & 2 \\
Art1 & artificial & 0.1 & 0.05 & 0.2 & 2 \\
Art2 & artificial & 0.2 & 0.1 & 0.4 & 2 \\
\end{tabular}
\end{subtable}
\begin{subtable}[c]{0.5\textwidth}
\centering
\subcaption{Test results}
\label{tab:rules_error}
\begin{tabular}{c|ccc}
test name & $\#$ & $L$ & $E$\\
\hline
Smokers1 & 2   & 3   & 0.004\\
Smokers2 & 2   & 4-7 & 0.31\\
Art1     & 1-2 & 1-4 & 0.01\\
Art2     & 1-2 & 1-4 & 0.021
\end{tabular}
\end{subtable}
\end{table}

\subsection{Distance and Reconstruction}
\begin{figure}[tb]
	\begin{subfigure}{.48\textwidth}
		\centering
		\includegraphics[width=.95\linewidth]{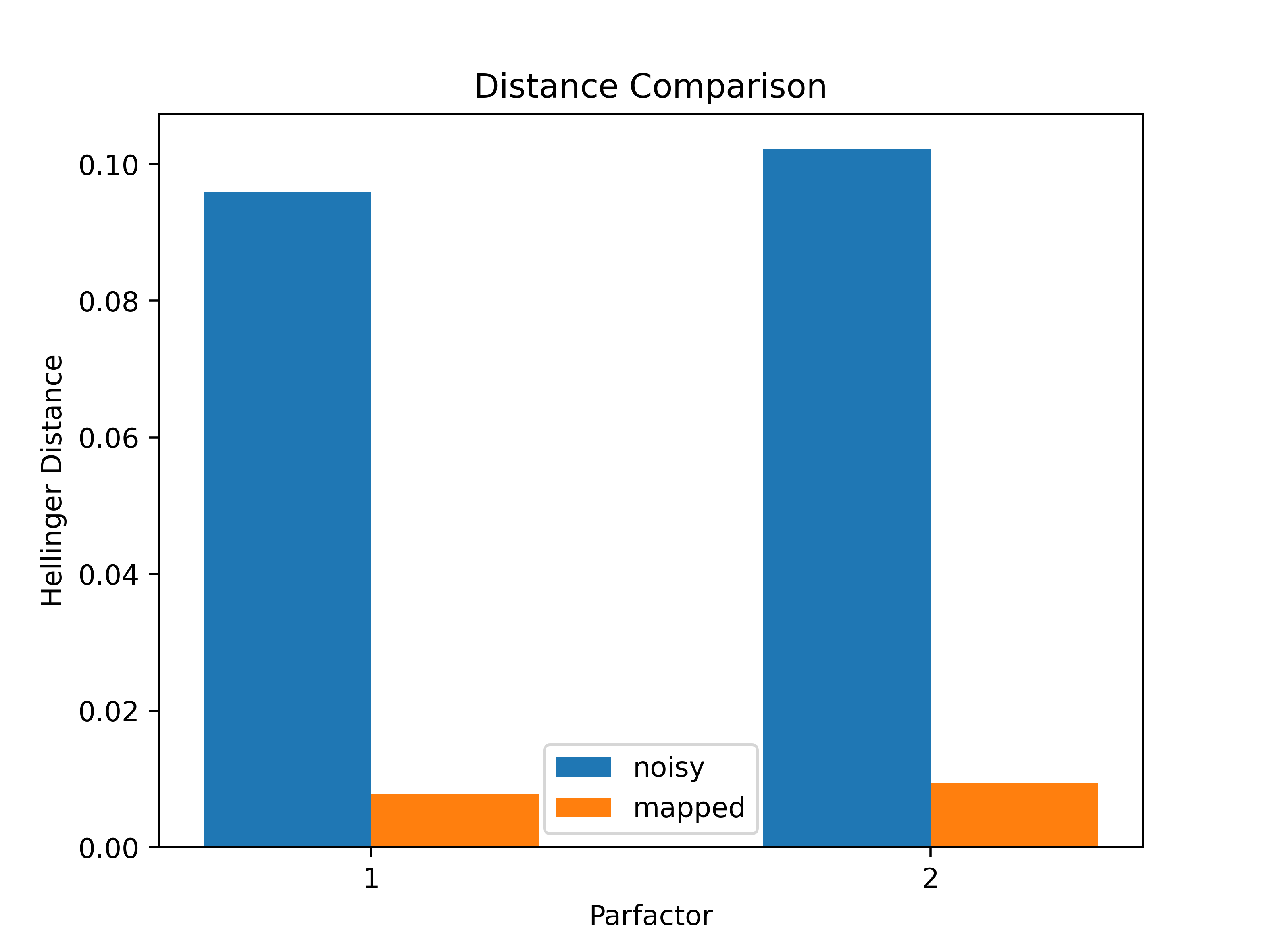}
		\caption{Smokers1}
		\label{fig:smokers_1_distances}
	\end{subfigure}%
	\hfill
	\begin{subfigure}{.48\textwidth}
		\centering
		\includegraphics[width=.95\linewidth]{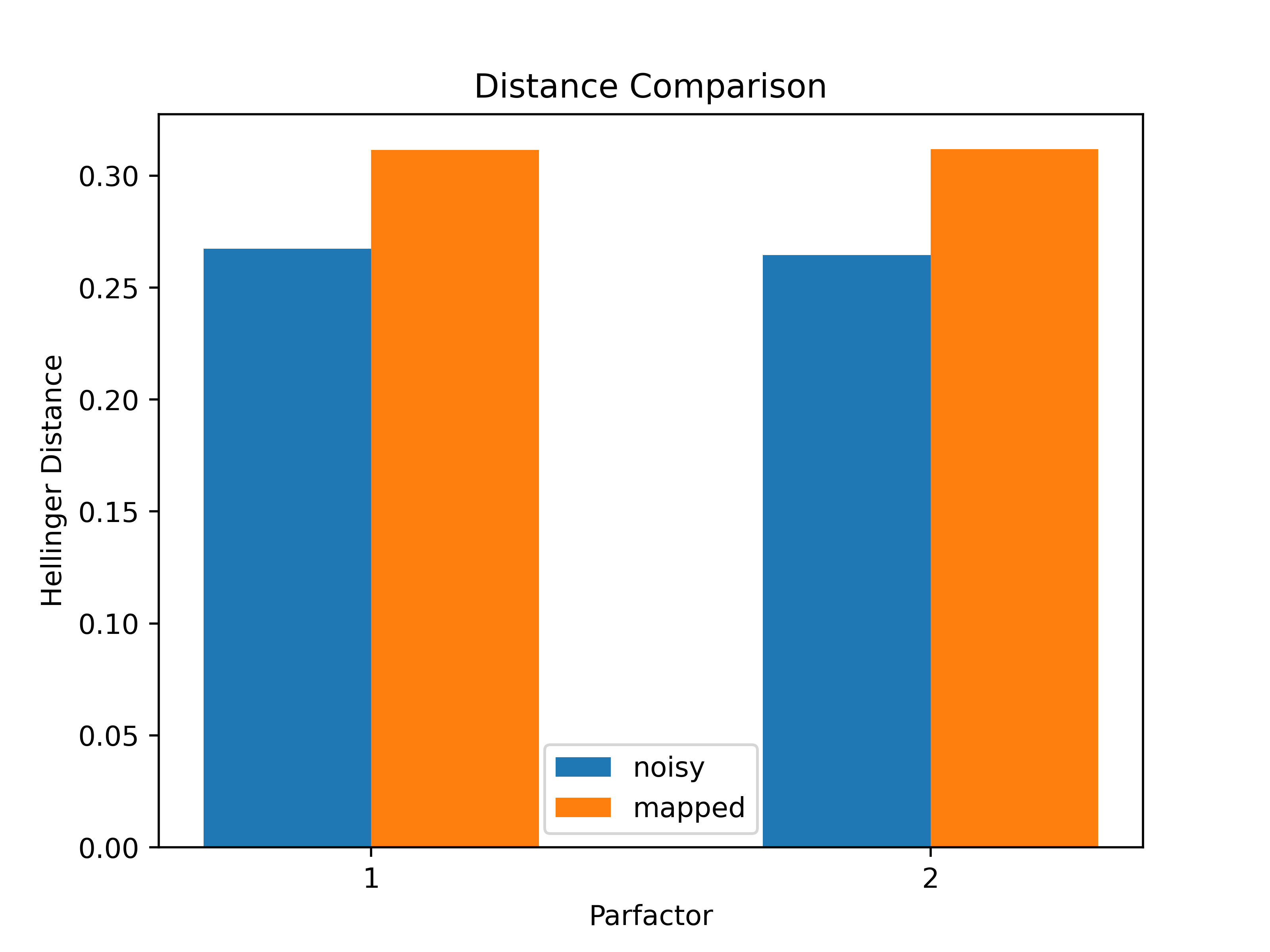}
		\caption{Smokers2}
		\label{fig:smokers_2_distances}
	\end{subfigure}
	\begin{subfigure}{.48\textwidth}
		\centering
		\includegraphics[width=.95\linewidth]{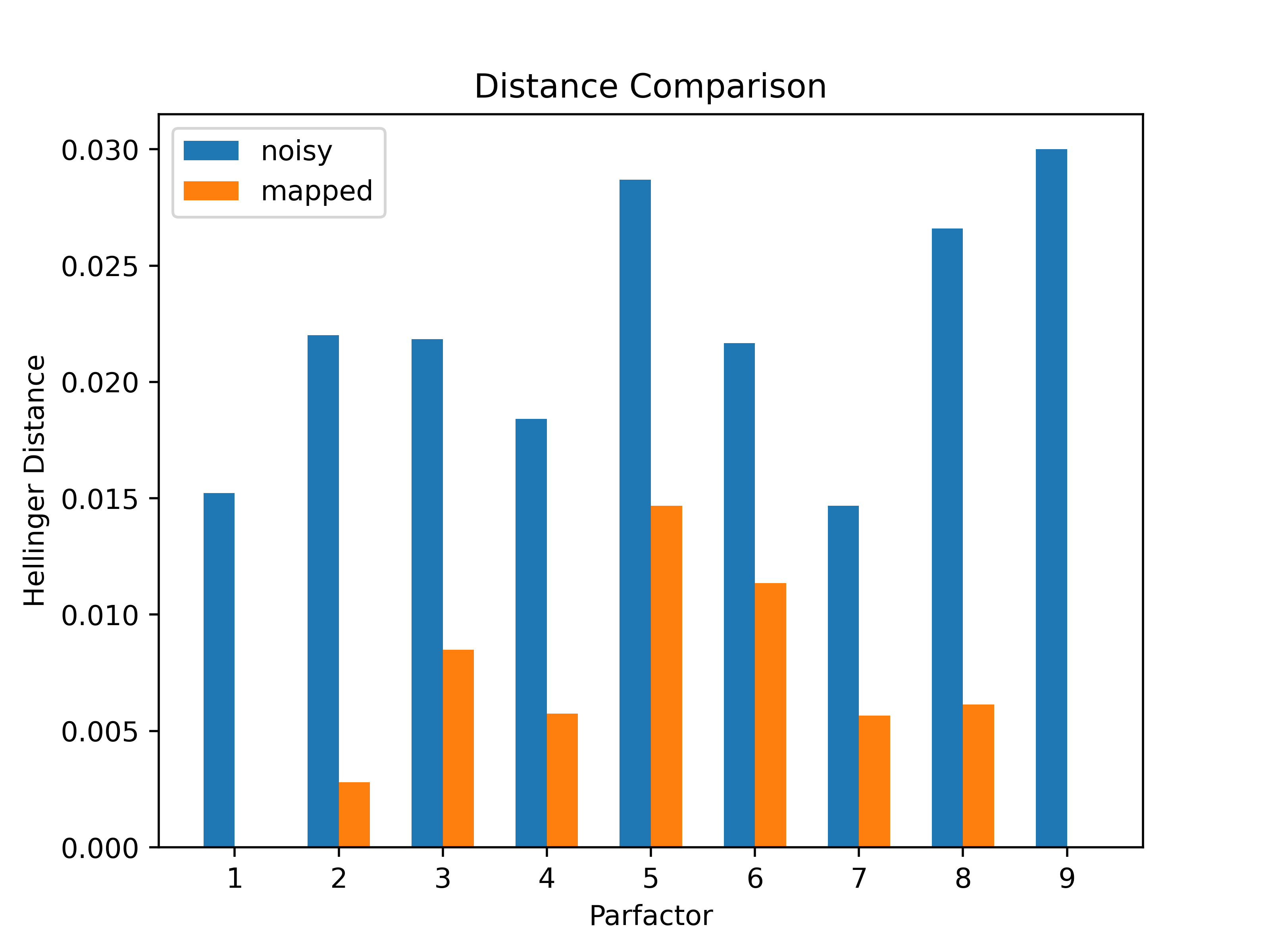}
		\caption{Art1}
		\label{fig:artificial_1_distances}
	\end{subfigure}%
	\hfill
	\begin{subfigure}{.48\textwidth}
		\centering
		\includegraphics[width=.95\linewidth]{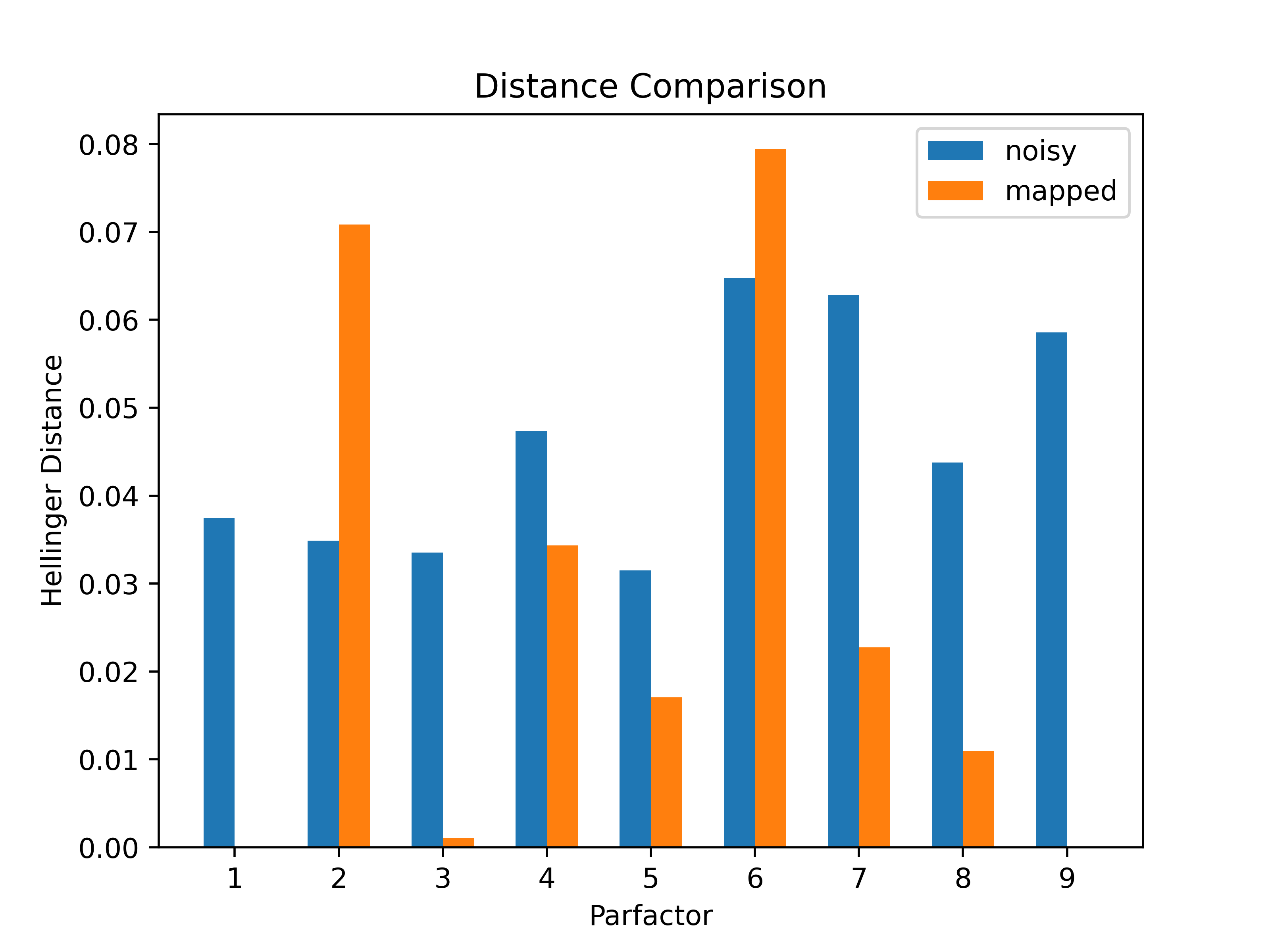}
		\caption{Art2}
		\label{fig:artificial_2_distances}
	\end{subfigure}
	\caption{Hellinger distances from the original to the noised and mapped models}
	\label{fig:distances}
\end{figure}

Due to adding noise to the potentials, the model used as input to CoFE has a certain Hellinger distance to the original model. \Cref{fig:distances} shows the Hellinger distances from the original model to the noised and mapped ones. For Smokers1 and Art1, CoFE can effectively filter out the noise and reconstruct the original distribution. For Smokers1, the mapped model has a Hellinger distance of $0.01$ to the original model. For Art1, CoFE at least halves the Hellinger distance compared to the one of the noised model. When we treat the original distribution as the latent one underlying the noised model in this tests, we capture the latent formulas.
Considering Smokers2, CoFE can no longer filter out the noise due to the high standard deviation compared to the absolute potentials. Moreover, CoFE not even approximately reconstructs the original distribution, but rather mixes the original clusters.
For most parfactors in Art2, CoFE can drastically reduce the noise and reconstruct the original clusters. But, CoFE mixes the two clusters in two parfactors. Thus, CoFE only roughly captures the original model.

\subsection{Sparsity and Error}
\Cref{tab:rules_error} shows the number and length of formulas extracted as well as the mean absolute error.
In all tests, we reduce the number of formulas exponentially compared to the canonical extraction. 
We extract two formulas per parfactor instead of the eight formulas we would get with the canonical transformation to an MLN. 
Moreover, all formulas, except for Smokers2, are only up to one atom longer than without simplification and minimization. 
For Smokers1, CoFE correctly returns the two formulas given in \cref{eq:smokers_rule1,eq:smokers_rule2}. 
Because of the added noise, the weights are slightly different: $0.074$ for \cref{eq:smokers_rule1} and $2.03$ for \cref{,eq:smokers_rule2}.

For Smokers1 and Art1, the error is small and at most $0.01$. 
Comparing Art2 to Art1, the error doubles as does the standard deviation of the noise added. 
For Smokers2, the mean absolute error is with $0.31$ clearly higher than for Smokers1 due to the higher noise standard deviation.

In summary, we can increase the sparsity of the encoding by extracting significantly less formulas than without reduction. 
Moreover, we preserve core information and the user can control how much information loss is tolerable.
Looking at the evaluation from a readability viewpoint, we hypothesize that a sparsely encoded distribution turned into an MLN formula is easier to understand or more readable as a human.
Additionally, the fewer formulas we have, the better we can understand these formulas; and the shorter the formula is, the better we can understand this particular formula.
Given this hypothesis, we can record that the formulas extracted with CoFE are at least as readable as the result of the canonical transformation from a parfactor to an MLN. 
In the worst case, our result would not differ in terms of readability from the one without reduction and minimization. 
With reduction, we may combine some formulas into one formula. 
Without minimization, this one formula is as long as the merged formulas together. 
With minimization, we can only shorten the formula or leaving it as it is.
Under this hypothesis, using CoFE on query answers enables another form of interpretation of query answers.
Future work includes to further investigate this avenue of readability, which also touches on ideas explored in transparent machine learning or explainable AI.

\section{Conclusion}
We present an algorithm to get an even more sparse encoding of a full joint distribution encoded in tabular-like parfactors.
To this end, the algorithm reduces the number of different potentials in a parfactor before extracting formulas, up to a user-defined maximum distance $\epsilon$ between the original distribution and the reduced distribution in the parfactor.
With the $\epsilon$, the user can trade off the potential for reduction with the preservation (accuracy) of the original distribution.
Specifically, we test out two different reduction strategies, based on clusters and quantiles respectively.
After formula extraction, a minimization step ensures that formulas are as short as possible.
Because of the reduction step, the algorithm can extract latent formulas hidden behind distorted potentials, while being able to preserve core information, with small errors observed in our evaluation with a small distance.
Furthermore, we hypothesize that human understanding can improve greatly by extracting few and short formulas. 

For future work, we look at testing out further reduction strategies, analyzing the relationship between the changes in the distribution and resulting reduction as well as error, and investigating the potential for increased readability and human understanding.

\bibliographystyle{splncs04}
\bibliography{references}

@inproceedings{GogDo11, 
  author = "Vibhav Gogate and Pedro Domingos",
  title = "{Probabilistic Theorem Proving}",
  booktitle = "UAI-11 Proc.\ of the 27th Conf.\ on Uncertainty in AI",
  pages = "256--265",
  year = "2011",
  publisher = {AUAI Press}
}

@article{NatKhKeGuSh12, 
  author = "Sriraam Natarajan and Tushar Khot and Kristian Kersting and Bernd Gutmann and Jude Shavlik",
  title = "{Gradient-based Boosting for Statistical Relational Learning: The Relational Dependency Network Case}",
  journal = "Machine Learning",
  pages = "25--56",
  volume = "86",
  year = "2012"
}

@article{AhmKeMlNa13, 
  author = "Babak Ahmadi and Kristian Kersting and Martin Mladenov and Sriraam Natarajan",
  title = "{Exploiting Symmetries for Scaling Loopy Belief Propagation and Relational Training}",
  journal = "Machine Learning",
  volume = "92",
  number = "1",
  pages = "91--132",
  year = "2013",
  publisher = {Springer}
}

@inproceedings{BraMo16a,
  author = {Tanya Braun and Ralf M\"oller}, 
  title = {{Lifted Junction Tree Algorithm}}, 
  booktitle = {Proc.\ of {KI} 2016: Advances in AI}, 
  pages = "30--42",
  year = {2016},
  publisher = "Springer"
}

@inproceedings{ChaDa07, 
  author = "Mark Chavira and Adnan Darwiche",
  title = "{Compiling Bayesian Networks Using Variable Elimination}",
  booktitle = "IJCAI-07 Proc.\ of the 20th International Joint Conf.\ on AI",
  pages = "2443--2449",
  year = "2007",
  publisher = {IJCAI Organization}
}

@article{mln,
  title="{Markov Logic Networks}",
  author={Richardson, Matthew and Domingos, Pedro},
  journal={Machine learning},
  volume={62},
  number={1-2},
  pages={107--136},
  year={2006},
  publisher={Springer}
}

@inproceedings{fo_inference,
   author = "David Poole",
  title = "{First-order Probabilistic Inference}",
  booktitle = "IJCAI-03 Proc.\ of the 18th International Joint Conf.\ on AI",
  pages = {985--991},
  year = "2003",
  publisher = {IJCAI Organization}
}

@ARTICLE{qmc,  author={McCluskey, E. J.},  journal={The Bell System Technical Journal},   title="{Minimization of Boolean Functions}",   year={1956},  volume={35},  number={6},  pages={1417-1444}}

@inproceedings{dbscan,
  title={{A Density-Based Algorithm for Discovering Clusters in Large Spatial Databases with Noise}},
  author={Ester, Martin and Kriegel, Hans-Peter and Sander, J{\"o}rg and Xu, Xiaowei},
  booktitle={KDD-96 Proc.\ of the 2nd International Conf.\ on Knowledge Discovery and Data Mining},
  pages={226--231},
  year={1996},
  organization = {AAAI Press}
}

@article{schubert2017dbscan,
  title={{DBSCAN Revisited, Revisited: Why and How You Should (Still) Use DBSCAN}},
  author={Schubert, Erich and Sander, J{\"o}rg and Ester, Martin and Kriegel, Hans Peter and Xu, Xiaowei},
  journal={ACM Transactions on Database Systems (TODS)},
  volume={42},
  number={3},
  pages={19},
  year={2017},
  publisher={ACM}
}

@article{parfactor_definitions,
  title="{Lifted Variable Elimination: Decoupling the Operators from the Constraint Language}",
  author={Taghipour, Nima and Fierens, Daan and Davis, Jesse and Blockeel, Hendrik},
  journal={Journal of Artificial Intelligence Research},
  volume={47},
  pages={393--439},
  year={2013}
}

@inproceedings{smokers,
  title="{Lifted Probabilistic Inference by First-order Knowledge Compilation}",
  author={Van den Broeck, Guy and Taghipour, Nima and Meert, Wannes and Davis, Jesse and De Raedt, Luc},
  booktitle={IJCAI-11 Proc.\ of the 22nd International Joint Conf.\ on AI},
  pages={2178--2185},
  year={2011},
  publisher = {IJCAI Organization}
}

@phdthesis{mln_to_parfactor, 
  author = "Guy {Van den Broeck}",
  title = "{Lifted Inference and Learning in Statistical Relational Models}",
  school = "KU Leuven",
  year = "2013"
}

@article{chakraborty2015generating,
  title={Generating discrete analogues of continuous probability distributions-A survey of methods and constructions},
  author={Chakraborty, Subrata},
  journal={Journal of Statistical Distributions and Applications},
  volume={2},
  number={1},
  pages={1--30},
  year={2015},
  publisher={Springer}
}

@inproceedings{RaeKiTo07,
  author={Luc {De Raedt} and Angelika Kimmig and Hannu Toivonen},
  title="{ProbLog: A Probabilistic Prolog and its Application in Link Discovery}",
  booktitle={IJCAI-07 Proceedings of 20th International Joint Conf.\ on AI},
  pages={2062--2467},
  year={2007},
  publisher = {IJCAI Organization}
}

\end{document}